\documentclass[letterpaper]{article} 
\usepackage[preprint]{aaai2027}
\usepackage[hyphens]{url}  
\usepackage{graphicx} 
\usepackage{natbib}  
\usepackage{caption} 
\usepackage{algorithm}
\usepackage{algorithmic}
\usepackage{amsmath}
\usepackage{amssymb}
\usepackage{multirow}

\newcommand{\tworowhead}[1]{%
  \multirow{2}{*}{\raisebox{-0.3em}{\textbf{#1}}}%
}
\usepackage{newfloat}
\usepackage{listings}
\DeclareCaptionStyle{ruled}{labelfont=normalfont,labelsep=colon,strut=off} 
\floatstyle{ruled}
\newfloat{listing}{tb}{lst}{}
\floatname{listing}{Listing}

\usepackage{booktabs}

\title{LEEPS: Latent-Guided Explore-Exploit Prompt Sampling for Efficient RLVR in Large Language Models}
\author{
    Shuang Liang\textsuperscript{\rm 1},
    Haoyang Zhou\textsuperscript{\rm 1},
    Yifan Gong\textsuperscript{\rm 1},
    Guowei Wang\textsuperscript{\rm 2},
    Xiting Wang\textsuperscript{\rm 1}\corresponding
}
\affiliations{
    \textsuperscript{\rm 1}Renmin University of China\\
    \textsuperscript{\rm 2}AntGroup\\
    xitingwang@ruc.edu.cn
}

\begin{document}

\maketitle

\begin{abstract}
Reinforcement learning with verifiable rewards (RLVR) improves the reasoning capabilities of large language models, but prompt groups with identical rollout rewards consume generation budget without effective learning signals. 
Pre-rollout prompt selection can reduce this waste by screening prompts before rollout generation. However, existing pre-rollout methods struggle to balance exploitation and exploration: repeatedly exploiting historically informative prompts can narrow training coverage, whereas broader exploration can lower the fraction of informative prompts.
To address these limitations, we introduce LEEPS, a Latent-Guided Explore--Exploit Prompt Sampler that adaptively balances the reuse of previously observed informative prompts with continued exploration of uncertain ones.
LEEPS partitions candidates into exploit and explore portfolios and adaptively allocates rollout budget according to their recent non-trivial ratios. It further uses representation-space neighbors and historical rollout outcomes to prioritize uncertain prompts likely to yield non-zero reward variance, thereby making exploration more targeted without additional rollouts. 
Across six mathematical reasoning benchmarks, LEEPS achieves the highest average score at both model scales, with relative gains of 2.6\% and 3.7\% over the strongest baseline for Qwen2.5-Math-1.5B and 7B, respectively, and generally improves faster during the training process. It also achieves the highest average score across the three evaluated OOD general-reasoning benchmarks at both model scales and adds only about 2 seconds of online sampling overhead per training step. Code is available at \url{https://github.com/ShuangLiangX/LEEPS}.
\end{abstract}

\section{Introduction}
Reinforcement learning has become a key post-training stage for improving the reasoning capabilities of Large Language Models (LLMs)~\cite{lambert2024tulu,guo2025deepseek,yu2026dapo}. Recent systems increasingly adopt outcome-level reinforcement learning with verifiable rewards (RLVR) for reasoning tasks~\cite{yu2026dapo,wen2025reinforcement,xie2025logic,hu2026open}, often optimized with Group Relative Policy Optimization (GRPO)~\cite{guo2025deepseek}. This paradigm enables models to learn from automatically checkable final answers, reducing reliance on preference-based reward modeling~\cite{ouyang2022training} and step-level process supervision~\cite{uesato2022solving,lightman2024let,wang2024math}. However, the combination of sparse outcome rewards and group-relative advantage estimation introduces a structural inefficiency: trivial (either too easy or too hard) prompts  often produce rollout groups with identical rewards, where all responses are either correct or incorrect~\cite{zheng2026act}. These zero-variance prompt groups yield vanishing signals for the model updates, yet still require expensive generation, making uninformative rollouts a major source of computational waste in RLVR training.

\begin{figure}[t]
  \centering
  \includegraphics[width=\columnwidth]{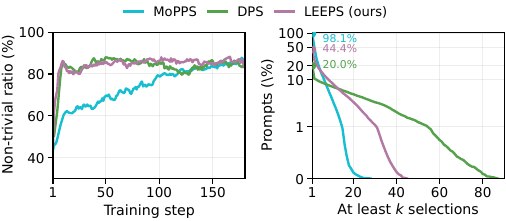}
\caption{Training dynamics on Qwen2.5-Math-7B. Left: the non-trivial prompt ratio over 180 training steps. Right: the complementary cumulative distribution of prompt selection counts at step 180, showing the fraction of training prompts selected at least $k$ times.}
  \label{fig:intro-motivation}
\end{figure}

Existing methods aim to increase the fraction of  non-trivial prompts  by selecting prompts. Online filtering methods adopt an observe-then-filter strategy: they first generate rollouts for sampled prompts and then discard zero-variance groups~\cite{yu2026dapo,zhang2025speed,xu2025not}. Despite their effectiveness, these methods still spend substantial computation on prompts that ultimately yield zero-variance groups. More recent methods move prompt selection before rollout by predicting prompt informativeness from rollout history. Specifically, DPS~\cite{mao2026dynamics} models prompt-solving dynamics to prioritize prompts likely to be informative, whereas MoPPS~\cite{qu2026can} uses posterior sampling over prompt success rates to favor prompts of intermediate predicted difficulty. Although pre-rollout selection avoids the additional rollouts required by online filtering, it introduces new challenges. As shown in Figure~\ref{fig:intro-motivation}, DPS maintains a high non-trivial ratio by repeatedly selecting a narrow prompt subset, leaving potentially useful training data unused, whereas MoPPS achieves broader coverage but yields a lower non-trivial ratio, particularly early in training. This setting gives rise to two core challenges. First, exploration itself can introduce zero-variance groups and thereby reduce the non-trivial ratio of the training batch. Second, under a fixed exploration budget, the sampler must identify potentially useful prompts more effectively without additional rollout feedback, thereby reducing the cost of exploration.

To address these challenges, we propose \textbf{LEEPS}, a latent-guided explore--exploit prompt sampler for efficient RLVR. As shown in Figure~\ref{fig:intro-motivation}, LEEPS maintains a consistently high non-trivial prompt ratio while covering substantially more prompts than DPS, thereby balancing training-batch informativeness with the exploration of potentially useful prompts. Specifically, LEEPS builds on two complementary designs: \textbf{Adaptive Explore--Exploit Portfolio Allocation} and \textbf{Latent-Guided Exploration}.

First, Adaptive Explore--Exploit Portfolio Allocation mitigates the reduction in the training-batch non-trivial ratio caused by exploration. It partitions candidates into an exploit portfolio of prompts that have produced informative rollout signals and an explore portfolio of prompts whose utility remains uncertain. Using recent rollout outcomes, LEEPS estimates the non-trivial ratio produced by each portfolio and adjusts their batch quotas so that the expected non-trivial ratio of the selected training batch remains close to a predefined target. This adaptive allocation maintains a high non-trivial ratio while preserving sufficient budget for exploration.

Second, Latent-Guided Exploration improves the efficiency of exploring uncertain prompts without additional rollouts. It represents prompts using model hidden states and estimates the potential success rate of each exploration candidate from the observed outcomes of its latent nearest neighbors. 
Figure~\ref{fig:latent-knn-auroc} shows that latent information provides a useful signal for screening uncertain prompts, supporting our latent-guided exploration strategy.
This allows the fixed exploration budget to prioritize uncertain prompts that are more likely to yield non-zero reward variance.

Finally, we evaluate LEEPS on two Qwen2.5-Math model scales across six mathematical reasoning benchmarks. Without requiring additional rollouts for prompt selection, LEEPS outperforms the strongest baseline in overall score by 2.6\% and 3.7\% for the 1.5B and 7B models, respectively. It also achieves the highest average score across three OOD general-reasoning benchmarks at both model scales, indicating that its gains extend beyond mathematical reasoning tasks. Further analyses show that LEEPS maintains a high non-trivial ratio, generally improves faster than the baselines, and incurs only approximately 2 seconds of online sampling overhead per training step.

\begin{figure*}[t]
    \centering
    \includegraphics[width=\textwidth]{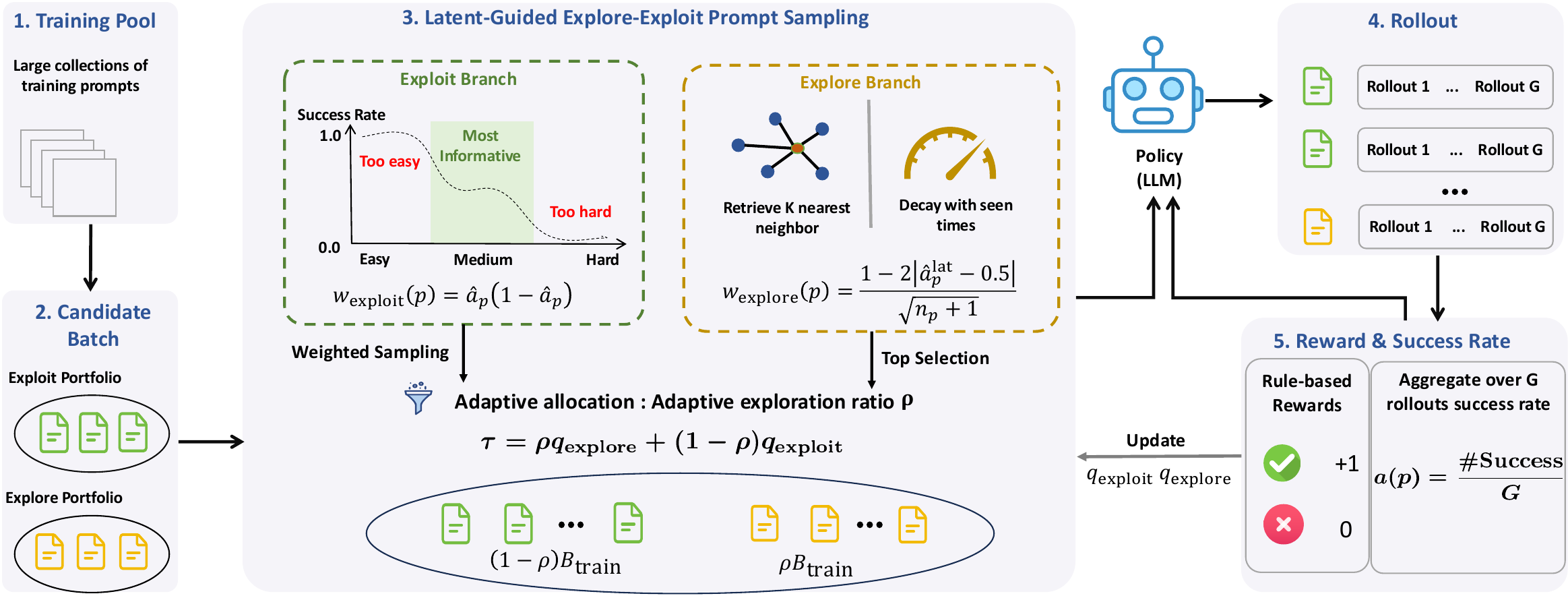}
    \caption{
    Overview of LEEPS. Candidate prompts are divided into exploit and explore portfolios, and the training budget is adaptively allocated according to their recent prompt non-trivial ratio. The exploit branch weights previously observed non-trivial prompts by their success rates, whereas the explore branch uses latent-neighbor estimates to identify and prioritize potentially informative prompts. Rollout outcomes update the prompt statistics for subsequent prompt sampling.
    }
    \label{fig:pipeline}
\end{figure*}

\section{Preliminaries}

\textbf{Zero-Variance Prompt Groups in RLVR.}
Group Relative Policy Optimization (GRPO)~\cite{shao2024deepseekmath} is widely used for reinforcement learning with verifiable rewards (RLVR). Given a prompt $p$, the old policy samples a group of $G$ responses $\{y_{p,j}\}_{j=1}^{G}$. We use a binary verifiable reward $R(p,y_{p,j})\in\{0,1\}$, where a correct final answer receives reward $1$ and an incorrect answer receives reward $0$. GRPO computes the group-relative advantage by normalizing each reward within its rollout group:
\begin{equation}
A_{p,j}
=
\frac{
R(p,y_{p,j})-\bar{R}_{p}
}{
\operatorname{std}(\mathbf{R}_{p})
},
\end{equation}

We summarize the rollout outcomes of prompt $p$ by its group success rate:
\begin{equation}
a(p)=\frac{1}{G}\sum_{j=1}^{G}R(p,y_{p,j}).
\end{equation}
When $a(p)=0$ or $a(p)=1$, all responses in the group receive the same reward, so $A_{p,j}=0$ for every response. We refer to the sampled rollout group as a \emph{zero-variance group} and regard $p$ as a \emph{trivial prompt} at that training step. In contrast, a prompt group is \emph{non-trivial} if $0<a(p)<1$, meaning that it contains both correct and incorrect responses. The non-trivial ratio of a prompt batch is the fraction of its prompt groups that satisfy this condition.

\noindent\textbf{Problem Definition.}
At each training step, the dataloader randomly samples a large candidate batch $\mathcal{B}_{\mathrm{cand}}$ with $|\mathcal{B}_{\mathrm{cand}}|>B$ following prior work~\cite{qu2026can,mao2026dynamics}, where $B$ is the number of prompts used for each GRPO update. Before generating rollouts, a pre-rollout selector uses only information available at selection time to choose a subset $\mathcal{B}_{\mathrm{train}}\subset\mathcal{B}_{\mathrm{cand}}$ with $|\mathcal{B}_{\mathrm{train}}|=B$. Only the selected prompts are rolled out and used for further policy optimization. The goal is to actively explore prompts whose utility remains uncertain while maintaining a high non-trivial ratio in the selected training batch, without generating additional rollouts for selection.


\section{Methods}

This section presents LEEPS, a latent-guided explore-exploit prompt sampler for efficient RLVR. As shown in Figure~\ref{fig:pipeline}, given a candidate batch $\mathcal{B}_{\mathrm{cand}}$, LEEPS selects a training batch $\mathcal{B}_{\mathrm{train}}$ through two components: Adaptive Explore--Exploit Portfolio Allocation, which distributes the rollout budget between the exploit and explore portfolios, and Latent-Guided Exploration, which uses latent-neighbor information to guide sampling within the explore portfolio. After the selected prompts are rolled out, LEEPS updates their success rates and selection counts using the observed outcomes, providing historical signals for the next selection step.

\subsection{Adaptive Explore--Exploit Portfolio Allocation}

To maintain a high non-trivial ratio without sacrificing broad exploration, we first introduce Adaptive Explore--Exploit Portfolio Allocation. As shown in Figure~\ref{fig:intro-motivation}, MoPPS~\cite{qu2026can} explores a broad portion of the training set but yields a lower non-trivial prompt-group ratio early in training. In contrast, DPS~\cite{mao2026dynamics} maintains a high non-trivial ratio by concentrating on prompts estimated to be useful, but consequently reuses a much smaller subset of prompts, leaving potentially non-trivial prompts unexplored. To balance the non-trivial ratio and prompt coverage, LEEPS formulates pre-rollout prompt selection as the adaptive allocation of a fixed rollout budget between two complementary prompt portfolios.

Given an enlarged candidate batch $\mathcal{B}_{\mathrm{cand}}$, LEEPS partitions the candidate prompts based on  their latest observed group success rates $\hat{a}_p$ and  the counts they have been selected for rollout, $n_p$. The exploit portfolio contains previously selected non-trivial prompts, whereas the explore portfolio contains unseen prompts and prompts whose latest rollout groups are zero-variance. LEEPS then constructs $\mathcal{B}_{\mathrm{train}}$ by adaptively allocating prompts from the two portfolios according to their recent non-trivial ratios.

\textbf{Exploit portfolio.}
The exploit portfolio contains prompts that have already produced non-zero reward variance:
\begin{equation}
\mathcal{P}_{\mathrm{exploit}} =
\{p \mid n_p > 0,\; 0 < \hat{a}_p < 1\}.
\end{equation}
These prompts are likely to provide useful learning signals because the current policy sometimes solves them and sometimes fails. LEEPS performs weighted sampling from the exploit portfolio, assigning each prompt a sampling probability proportional to its Bernoulli-variance weight:
\begin{equation}
w_{\mathrm{exploit}}(p) = \hat{a}_p(1-\hat{a}_p),
\end{equation}
which is maximized at $\hat{a}_p=0.5$ and decreases as the prompt becomes either easier or harder. This weighting therefore favors prompts of intermediate difficulty, consistent with prior findings that these prompts  are most informative for model learning~\cite{bae2026online,chen2025self,qu2026can}.

\textbf{Explore portfolio.}
The explore portfolio contains prompts whose utility remains uncertain under the current policy. It includes both unseen prompts and prompts whose latest rollout outcomes are zero-variance:
\begin{equation}
\mathcal{P}_{\mathrm{explore}} =
\{p \mid n_p = 0\} \cup
\{p \mid n_p > 0,\; \hat{a}_p \in \{0,1\}\}.
\end{equation}
Keeping these prompts in the explore portfolio prevents the sampler from prematurely discarding prompts after limited or uninformative observations. This is important because a prompt that is zero-variance at one stage of training may later become useful as the policy evolves. LEEPS does not sample this portfolio randomly; instead, it prioritizes exploring candidates using a latent-neighbor estimate of their potential success rate, which is detailed in the next subsection.
Let $\hat{a}^{\mathrm{lat}}_p$ denote this latent-neighbor estimated success rate. LEEPS assigns each explore prompt a count-decayed latent uncertainty score and selects the highest-scoring prompts:
\begin{equation}
\label{eq:explore-score}
w_{\mathrm{explore}}(p) =
\frac{
 1 - 2|\hat{a}^{\mathrm{lat}}_p - 0.5|
}{
\sqrt{n_p+1}
},
\end{equation}
which favors prompts whose latent-neighbor estimate suggests intermediate difficulty, and discourages repeatedly exploring the same prompts.  

\textbf{Adaptive allocation.}
LEEPS aims to keep the non-trivial ratio of the selected training batch close to a predefined target $\tau$ (e.g., $0.9$). To achieve this target, LEEPS controls the explore--exploit allocation through $\rho$, the fraction of the batch assigned to the explore portfolio. Specifically, approximately $\rho B$ prompts are selected from the explore portfolio and $(1-\rho)B$ from the exploit portfolio to form $\mathcal{B}_{\mathrm{train}}$.
To determine $\rho$, LEEPS uses the recent non-trivial ratios of prompts selected from the two portfolios, denoted by $q_{\mathrm{explore}}$ and $q_{\mathrm{exploit}}$. The estimated non-trivial ratio of the resulting training batch is
\begin{equation}
\rho q_{\mathrm{explore}}
+
(1-\rho)q_{\mathrm{exploit}},
\end{equation}
and LEEPS chooses $\rho$ such that this estimate is as close as possible to the target $\tau$. After the selected prompts are rolled out, their outcomes are used to update $q_{\mathrm{explore}}$ and $q_{\mathrm{exploit}}$, which  determine the allocation ratio for the next training step. For stability, $\rho$ is clipped to a predefined range.

\subsection{Latent-Guided Exploration}

To improve the efficiency of exploring prompts with uncertain utility, LEEPS introduces Latent-Guided Exploration. Unseen and zero-variance prompts provide limited direct evidence about whether they will yield useful reward variation in subsequent rollouts, while random exploration treats all such prompts equally. 
Prior work has primarily examined whether question-only model representations can predict single-response correctness under fixed inference settings~\cite{cencerrado2025no,zhang2025self}. LEEPS instead focuses on identifying prompts likely to yield non-trivial rollout groups as the policy evolves.
Specifically, LEEPS extends prompt-level prediction to local representation-space neighborhoods, using the observed rollout behavior of nearby prompts to guide the exploration of uncertain candidates without additional rollouts.

\begin{figure}[t]
\centering
\includegraphics[width=\linewidth]{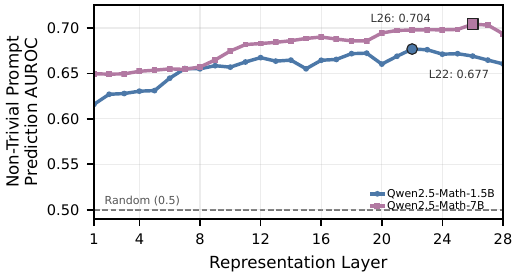}
\caption{Layer-wise AUROC of the latent-neighbor score across contextual hidden layers for predicting whether prompts are non-trivial on DAPO-Math-17K. The dashed line marks random prediction (AUROC $=0.5$).}
\label{fig:latent-knn-auroc}
\end{figure}

Figure~\ref{fig:latent-knn-auroc} supports this design: representing prompts with model hidden states and aggregating the success rates of their nearest neighbors provides a useful signal for identifying prompts likely to yield non-trivial rollout groups. Specifically, the figure evaluates the latent uncertainty term $1-2|\hat{a}^{\mathrm{lat}}_p-0.5|$ in Equation~\ref{eq:explore-score} without selection-count decay. This signal generally becomes stronger at deeper layers and reaches its best performance near the upper layers, suggesting that latent neighborhoods capture information relevant to group-level rollout outcomes. 
Based on this observation, LEEPS retrieves neighbors with observed rollout outcomes and aggregates their success rates to obtain the latent-neighbor estimate $\hat{a}^{\mathrm{lat}}_p$, which is then used to prioritize candidates within the explore portfolio.


  \begin{table*}[t]
  \centering
  \begin{tabular*}{\textwidth}{@{\extracolsep{\fill}} c l c c c c c c c c c @{}}
  \toprule
  \tworowhead{Model} &
  \tworowhead{Method} &
  \multicolumn{3}{c}{\textbf{Pass@1}} &
  \multicolumn{3}{c}{\textbf{Avg@16}} &
  \tworowhead{Avg.} &
  \tworowhead{Rollouts $\downarrow$} &
  \tworowhead{Time (h) $\downarrow$} \\
  \cmidrule(lr){3-5}\cmidrule(lr){6-8}
  & & \textbf{MATH} & \textbf{Miner.} & \textbf{Olym.}
  & \textbf{AMC23} & \textbf{AIME24} & \textbf{AIME25} & & & \\
  \midrule
   & Base & 37.20 & 11.40 & 21.69 & 30.31 & 6.88 & 4.17 & 18.61 & -- & -- \\
   & GRPO & \underline{75.00} & 29.04 & 37.35 & 55.62 & 15.42 & 9.79 & 37.04 & \textbf{573K} & \ \textbf{6.8} \\
   & DS & 74.20 & 29.41 & \textbf{39.07} & 58.44 & 15.42 & 11.67 & 38.03 & \underline{1260K} & 10.4 \\
  \smash{\shortstack{Qwen2.5-Math\\1.5B}} & MoPPS & \textbf{76.20} & \underline{30.88} & 37.35 & 56.25 & 15.21 & \underline{11.88} & 37.96 & \textbf{573K} & \underline{7.0} \\
   & DPS & 73.60 & \textbf{31.25} & 36.32 & \textbf{62.03} & \underline{15.62} & 10.21 & \underline{38.17} & \textbf{573K} & \underline{7.0} \\
   & LEEPS & \underline{75.00} & 29.78 & \underline{38.73} & \underline{60.62} & \textbf{18.54} & \textbf{12.29} & \textbf{39.16} & \textbf{573K} & \textbf{6.8} \\
  \midrule
   & Base & 52.20 & 16.91 & 17.90 & 33.75 & 10.83 & 4.58 & 22.70 & -- & -- \\
   & GRPO & 80.20 & 36.76 & 44.41 & 66.88 & 27.29 & 15.62 & 45.19 & \textbf{369K} & \textbf{12.1} \\
   & DS & 80.80 & 34.19 & \textbf{45.27} & 66.41 & \underline{32.08} & 14.58 & 45.56 & \underline{768K} & 16.4 \\
  \smash{\shortstack{Qwen2.5-Math\\7B}} & MoPPS & \textbf{82.00} & 36.03 & 43.55 & 67.81 & 29.79 & 15.42 & 45.77 & \textbf{369K} & \underline{12.7} \\
   & DPS & 79.80 & \underline{37.13} & \underline{44.92} & \underline{68.12} & 29.17 & \underline{15.83} & \underline{45.83} & \textbf{369K} & \underline{12.7} \\
   & LEEPS & \underline{81.60} & \textbf{38.97} & 43.55 & \textbf{71.09} & \textbf{33.96} & \textbf{16.04} & \textbf{47.53} & \textbf{369K} & \textbf{12.1} \\
  \bottomrule
  \end{tabular*}
\caption{Main results on mathematical reasoning benchmarks. The Rollouts column reports the cumulative number of responses generated during RL training, including those from zero-variance groups that are subsequently discarded by online filtering. Best and second-best results within each model group are shown in \textbf{bold} and \underline{underlined}, respectively.}
  \label{tab:main-results}
  \end{table*}

\textbf{Latent prompt representation.}
For each prompt $p$, LEEPS extracts the hidden state of its final prompt token from a selected model layer and applies L2 normalization to obtain the representation $h_p$. Before RL training, LEEPS precomputes a static $K$-nearest-neighbor cache over the full training set using cosine similarity:
\begin{equation}
c(p,p') = h_p^\top h_{p'}.
\end{equation}

\textbf{Neighbor-based success-rate estimation.}
For each explore candidate $p$, LEEPS estimates its potential success rate from neighbors with observed rollout outcomes. Let $\mathcal{N}_{K}^{\mathrm{obs}}(p)$ denote the subset of its $K$ nearest neighbors with available success rates. LEEPS computes
\begin{equation}
\hat{a}^{\mathrm{lat}}_p =
\frac{
\sum_{p' \in \mathcal{N}_{K}^{\mathrm{obs}}(p)}
\max(c(p,p'),0)\hat{a}_{p'}
}{
\sum_{p' \in \mathcal{N}_{K}^{\mathrm{obs}}(p)}
\max(c(p,p'),0)
},
\end{equation}
where $\max(c(p,p'),0)$ ensures non-negative similarity weights. The resulting estimate is used in $w_{\mathrm{explore}}(p)$ to guide candidate selection.

\section{Experiments}
In this section, we evaluate LEEPS to demonstrate its effectiveness, the necessity of its key components, and its computational efficiency. Additional implementation details and experiments are provided in the Appendix.

\subsection{Experimental Setup}
\textbf{Models and Training Datasets.}
We conduct experiments on two widely used backbone models, Qwen2.5-Math-1.5B and Qwen2.5-Math-7B~\cite{yang2024qwen25mathtechnicalreportmathematical}. All methods are trained on the same DAPO-Math-17K~\cite{yu2026dapo} training set to ensure a fair comparison. We focus on mathematical reasoning and use the rule-based verifier for both RLVR training and in-domain evaluation. Our experiments are implemented using VERL~\cite{sheng2025hybridflow} and vLLM~\cite{kwon2023efficient}. The 1.5B models are trained on 8 NVIDIA H20 GPUs, whereas the 7B models are trained on 8 NVIDIA A100 GPUs. Following prior work~\cite{qu2026can,mao2026dynamics}, all methods use a prompt batch size of 256 and sample 8 rollouts per prompt. We set the maximum prompt and response lengths to 1024 and 3072 tokens.

\noindent \textbf{Baselines.}
To evaluate LEEPS, we compare it against four representative baselines: (1) vanilla GRPO~\cite{shao2024deepseekmath}, which uniformly learns from the training data without prompt selection; (2) DS~\cite{yu2026dapo}, an online filtering method that filters prompt groups after rollout; 
(3) MoPPS~\cite{qu2026can}, a pre-rollout method that uses posterior sampling over prompt success rates to favor prompts of intermediate predicted difficulty; and (4) DPS~\cite{mao2026dynamics}, a pre-rollout method that models prompt-solving dynamics to prioritize informative prompts.

\noindent \textbf{Evaluation Benchmark.} We evaluate models on MATH-500~\cite{hendrycks2021measuring}, Minerva-Math~\cite{lewkowycz2022solving}, OlympiadBench~\cite{he2024olympiadbench}, AMC23, AIME24, and AIME25~\cite{li2024numinamath}. Following our evaluation protocol, we report pass@1 on MATH-500, Minerva-Math, and OlympiadBench. Because AMC23, AIME24, and AIME25 contain relatively few and challenging problems, we report avg@16 by sampling 16 responses per problem with temperature 0.6 and top-p 1.0.  To examine out-of-distribution (OOD) generalization, we additionally evaluate the MATH-trained models on GPQA-Diamond~\cite{rein2023gpqa}, ARC-C~\cite{allenai:arc}, and MMLU-Pro~\cite{wang2024mmlu}, reporting pass@1 accuracy.

\begin{figure*}[t]
\centering
\includegraphics[width=\textwidth]{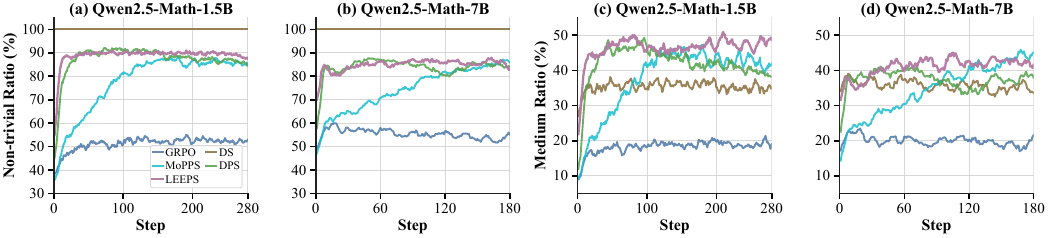}
\caption{Training-sample characteristics for Qwen2.5-Math-1.5B and Qwen2.5-Math-7B. Panels (a)--(b) report the non-trivial ratio, and Panels (c)--(d) report the medium ratio, defined as the fraction of selected prompt groups with $0.3 \le a(p) \le 0.7$. Curves are smoothed using a 7-step moving average.}
\label{fig:sample-characteristics}
\end{figure*}

\begin{figure*}[t]
\centering
\includegraphics[width=\textwidth]{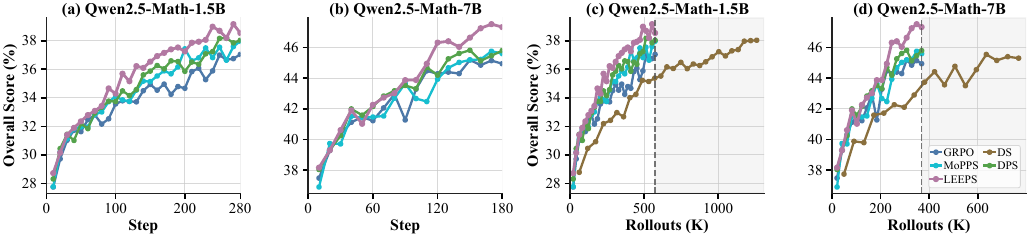}
\caption{Training progress for 1.5B and 7B models. Panels (a)--(b) show overall score versus training step, and Panels (c)--(d) versus cumulative rollouts, with DS included only in the latter. Dashed lines mark the shared rollout budgets of GRPO and the pre-rollout selectors; shaded regions indicate DS rollouts beyond those budgets.}
\label{fig:training-progress}
\end{figure*}

\subsection{Main Results}

This subsection evaluates LEEPS using benchmark results and complementary training diagnostics. We examine three aspects: overall performance, the characteristics of the selected training samples, and training progress.

\textbf{Overall Performance.} As summarized in Table~\ref{tab:main-results}, LEEPS achieves the highest overall score at both model scales. It outperforms DPS, the strongest baseline at both scales, by 0.99 and 1.70 points for the 1.5B and 7B models, corresponding to relative improvements of 2.6\% and 3.7\%, respectively, without requiring additional rollouts.

\textbf{Training Sample Characteristics.} LEEPS maintains a high non-trivial ratio while selecting a larger proportion of prompts with intermediate difficulty.
Figure~\ref{fig:sample-characteristics} compares the non-trivial ratio and the medium ratio, defined as the fraction of selected prompt groups whose success rates satisfy $0.3 \le a(p) \le 0.7$. Uniform GRPO produces  the lowest non-trivial ratios at both model scales. DS maintains the highest non-trivial ratio through post-rollout filtering but, as shown in Table~\ref{tab:main-results}, requires a substantially larger rollout budget. Moreover, despite its high non-trivial ratio, DS retains a smaller proportion of medium-difficulty prompt groups, which tend to provide more informative learning signals. This smaller proportion may partly explain why DS achieves lower final overall scores than the pre-rollout selectors despite its high non-trivial ratio.
Among the pre-rollout selection methods, DPS maintains a high non-trivial ratio but selects a smaller proportion of medium-difficulty prompts, whereas MoPPS eventually reaches a comparable medium ratio but exhibits a substantially lower non-trivial ratio in the early training stage. 
LEEPS combines a consistently high non-trivial ratio with a large proportion of medium-difficulty prompts throughout training.

\textbf{Training Progress.}
LEEPS reaches the highest overall score at both model scales and generally improves faster than the baselines at the same training step, as shown in Figure~\ref{fig:training-progress}(a)--(b). This advantage persists in the rollout-normalized comparison in Figure~\ref{fig:training-progress}(c)--(d). Within the common rollout budget, LEEPS outperforms GRPO and the other pre-rollout selectors, whereas DS remains below these methods and requires substantially more rollouts to complete training. Because DS may generate multiple candidate batches for each policy update, it is included only in the rollout-normalized panels.
Figure~\ref{fig:intro-motivation} further shows that LEEPS covers substantially more prompts than DPS while maintaining a higher non-trivial ratio than MoPPS, particularly early in training. The performance gains of LEEPS are therefore associated with a high non-trivial ratio, informative sample selection, and broad prompt coverage.

\subsection{Ablation Study}

\begin{figure}[htbp]
\centering
\includegraphics[width=\linewidth]{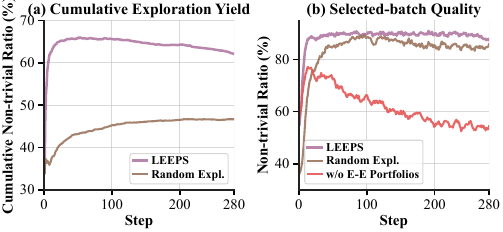}
\caption{Ablation diagnostics on Qwen2.5-Math-1.5B. Panel (a) reports the cumulative non-trivial ratio among explored prompt groups. Panel (b) compares the selected-batch non-trivial ratio of LEEPS and its ablations, with curves smoothed using a 7-step moving average.}
\label{fig:ablation-diagnostics}
\end{figure}

We conduct ablation studies to examine how the components contribute to LEEPS. As shown in Table~\ref{tab:ablation}, the complete method achieves the best overall score at both model scales. Figure~\ref{fig:ablation-diagnostics} further provides analysis for the random-exploration and no-portfolio variants on the 1.5B model. Specifically, we consider the following controlled variants:
\begin{itemize}
    \item \textit{w/ Random Exploration}: uniformly samples from the explore portfolio without Latent-Guided Exploration.
    \item \textit{w/ Uniform Exploitation}: uniformly samples from the exploit portfolio without success-rate-based weights.
    \item \textit{w/o E-E Portfolios}: removes Adaptive Explore--Exploit Portfolio Allocation and uses only Latent-Guided Exploration for prompt selection.
\end{itemize}

\begin{table}[htbp]
\centering
\begin{tabular*}{\linewidth}{@{\extracolsep{\fill}} lcc @{}}
\toprule
\textbf{Variant} & \textbf{1.5B} & \textbf{7B} \\
\midrule
\shortstack[l]{w/ Random Exploration} & 37.51 & 45.86 \\
\shortstack[l]{w/ Uniform Exploitation} & 38.48 & 46.29 \\
\shortstack[l]{w/o E-E Portfolios} & 37.94 & 45.27 \\
LEEPS & \textbf{39.16} & \textbf{47.53} \\
\bottomrule
\end{tabular*}
\caption{Ablation results for LEEPS. Values are the best overall scores within the model-specific selection ranges, across six mathematical reasoning benchmarks.}
\label{tab:ablation}
\end{table}

\textbf{Random Exploration.}
Replacing latent-guided exploration with random sampling results in relative overall score decreases of 4.2\% and 3.5\% for the 1.5B and 7B models, respectively. Because prompts in the explore portfolio are unseen or have produced zero-variance groups, random sampling treats candidates with substantially different potential utility equally. Figure~\ref{fig:ablation-diagnostics}(a) shows a lower cumulative non-trivial ratio among explored prompt groups, while Figure~\ref{fig:ablation-diagnostics}(b) shows a lower selected-batch non-trivial ratio. These results indicate that latent information transfers observations from related prompts to make exploration more targeted and maintain a more reliable supply of useful training signals.

\textbf{Uniform Exploitation.}
Uniform exploit sampling results in relative overall score decreases of 1.7\% and 2.6\% for the 1.5B and 7B models, respectively. Because this variant retains the portfolio structure and latent-guided exploration, the drop isolates the benefit of success-rate-aware weighting, which prioritizes informative prompts near the current learning boundary. Despite these reductions, the variant still outperforms the strongest baseline at both model scales, while full LEEPS achieves a larger margin.

\textbf{Without Explore-Exploit Portfolio.}
Removing the explore--exploit portfolios results in relative overall score decreases of 3.1\% and 4.8\% for the 1.5B and 7B models, respectively. Without separate portfolios, the sampler cannot explicitly retain verified useful prompts while reserving budget for continued exploration. Figure~\ref{fig:ablation-diagnostics}(b) shows that selected-batch quality consequently deteriorates during training; the proportion of all-one prompt groups rises from 6.3\% over the first 50 steps to 21.9\% over the final 50 steps. This result indicates that latent-guided exploration can prioritize uncertain candidates but cannot alone maintain a stable supply of informative prompts as the policy evolves.

\subsection{Additional Analysis}

\textbf{OOD Generalization.}
To assess whether improved in-domain training efficiency compromises broader reasoning ability, we evaluate the trained models on three OOD general-reasoning benchmarks. LEEPS achieves the highest average OOD score at both model scales, reaching 28.08 for the 1.5B model and 40.95 for the 7B model, compared with 27.94 and 40.85 for the strongest respective baselines (Table~\ref{tab:ood-results}). Although the margins are modest, these consistent gains across both model scales indicate that LEEPS improves in-domain training efficiency without sacrificing OOD general-reasoning performance.

\begin{table}[htbp]
\centering
\begin{tabular*}{\linewidth}{@{\extracolsep{\fill}} l l c c c c @{}}
\toprule
\textbf{Model} & \textbf{Method} & \textbf{GPQA} & \textbf{ARC-c} & \textbf{MMLU-P} & \textbf{Avg.} \\
\midrule
1.5B & Base & 5.56 & 2.65 & 4.62 & 4.27 \\
 & GRPO & \textbf{14.65} & 50.94 & 15.33 & 26.97 \\
 & DS & \textbf{14.65} & 53.67 & 15.50 & \underline{27.94} \\
 & MoPPS & 10.10 & \textbf{54.61} & \textbf{16.35} & 27.02 \\
 & DPS & 13.64 & 52.65 & 15.54 & 27.27 \\
 & LEEPS & \underline{14.14} & \underline{54.01} & \underline{16.07} & \textbf{28.08} \\
\midrule
7B & Base & 10.61 & 10.58 & 10.31 & 10.50 \\
 & GRPO & 18.69 & \underline{75.68} & 26.01 & 40.13 \\
 & DS & 20.20 & 75.09 & \underline{26.10} & 40.46 \\
 & MoPPS & \underline{20.71} & 75.00 & \textbf{26.30} & 40.67 \\
 & DPS & \underline{20.71} & \textbf{75.77} & 26.08 & \underline{40.85} \\
 & LEEPS & \textbf{21.21} & \underline{75.68} & 25.95 & \textbf{40.95} \\
\bottomrule
\end{tabular*}
\caption{OOD general reasoning results. Best and second-best results within each model group are shown in \textbf{bold} and \underline{underlined}, respectively.}
\label{tab:ood-results}
\end{table}

\begin{figure}[htbp]
\centering
\includegraphics[width=\linewidth]{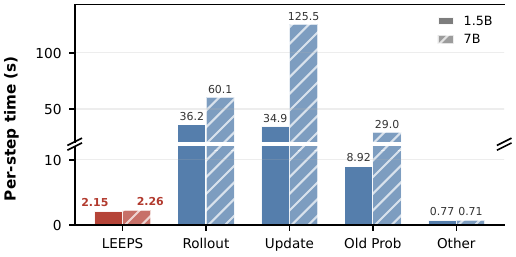}
\caption{Average per-step runtime over 20 profiled steps for both model scales on 8 NVIDIA H20 GPUs. LEEPS includes prompt selection and sampler updates.}
\label{fig:time-breakdown}
\end{figure}

\noindent \textbf{Computational Efficiency.}
Figure~\ref{fig:time-breakdown} shows that rollout generation and policy updates dominate runtime, while prompt selection and sampler updates add only 2.15 seconds (2.6\%) and 2.26 seconds (1.0\%) per step for the 1.5B and 7B models, respectively. Sampler initialization takes 124 and 292 seconds, respectively, but occurs only once and does not affect steady-state efficiency. The end-to-end runtimes reported in Table~\ref{tab:main-results} also show the efficiency of LEEPS.


\section{Related Work}
\noindent\textbf{RL for LLM Training.}
Reinforcement learning has been widely used in the post-training of large language models to align model outputs with human preferences and task-specific objectives. A representative line of work is RLHF, which learns reward models from human preference annotations and optimizes language models with PPO~\cite{schulman2017proximal,ouyang2022training,bai2022training}. Later preference-optimization methods, such as DPO, simplify this pipeline by directly optimizing preference pairs without explicit reward-model training or online RL updates~\cite{rafailov2023direct}. More recent work explores reinforcement learning with verifiable rewards, where automatically checkable signals in domains such as mathematics and code are used to supervise LLM reasoning~\cite{cobbe2021training,guo2025deepseek}. On the algorithmic side, GRPO further reduces the complexity of PPO-style optimization by using group-relative rewards instead of a separate value model~\cite{shao2024deepseekmath}. Building on this line of work, we address the inefficiency caused by zero-variance prompt groups in RLVR through pre-rollout prompt selection.

\noindent\textbf{Data Selection for RLVR.}
Prompt selection is important for RLVR because prompt utility changes throughout training. Traditional curriculum-based methods organize prompts using predefined difficulty levels or adjust the proportions of tasks at different difficulty levels~\cite{team2025kimi,song2025fastcurl,parashar2025curriculum}, but such signals may not capture the model's evolving competence. To obtain model-specific signals, online filtering methods use fresh rollouts to identify informative prompts~\cite{bae2026online,yu2026dapo,zhang2025speed,xu2025not}, incurring generation cost before filtering. Building on this paradigm, history-aware methods reuse past rollout accuracy as prior information to guide subsequent sampling and reduce redundant rollouts~\cite{chen2025self,zheng2026act,shen2025bots,shi2025efficient}. Complementary directions improve rollout efficiency by adaptively allocating different numbers of responses to prompts according to their estimated learning value~\cite{yao2026optimizing,zeng2025cures,fang2026allocate}, or by training auxiliary models to predict prompt difficulty~\cite{gao2025prompt,sun2026improving}. More recent methods make prompt-selection decisions without additional rollouts using historical training signals~\cite{qu2026can,mao2026dynamics}. However, existing pre-rollout selectors face an exploration--exploitation trade-off between exploiting historically informative prompts and exploring prompts with uncertain utility. LEEPS addresses this trade-off through adaptive explore--exploit portfolio allocation and latent-guided exploration.

\section{Conclusion}
We presented LEEPS, a pre-rollout prompt sampler that combines adaptive explore--exploit portfolio allocation with latent-guided exploration to select informative prompts and use fixed rollout budgets more effectively without requiring additional selection rollouts. Empirical results across six mathematical reasoning benchmarks demonstrate that LEEPS achieves the highest overall scores for both Qwen2.5-Math-1.5B and 7B under matched rollout budgets, and ablation studies further support the contributions of both components. LEEPS also achieves the highest average score across the three evaluated OOD benchmarks at both model scales, while incurring only minor online sampling overhead.

\bibliography{aaai2027}
\clearpage
\appendix

\section{Additional Method Details}
\label{appendix:method-details}

\subsection{GRPO Objective and Zero-Variance Handling}
\label{appendix:grpo}

Given a prompt $p$, the old policy samples a group of $G$ responses:
\begin{equation}
\mathcal{Y}(p)=\{y_{p,j}\}_{j=1}^{G},
\qquad
y_{p,j}\sim\pi_{\theta_{\mathrm{old}}}(\cdot\mid p).
\end{equation}
Let $\mathbf{R}_{p}=[R(p,y_{p,1}),\ldots,R(p,y_{p,G})]$ and let $\bar{R}_{p}$ denote its mean. GRPO computes the group-normalized advantage as
\begin{equation}
A_{p,j}
=
\frac{
R(p,y_{p,j})-\bar{R}_{p}
}{
\operatorname{std}(\mathbf{R}_{p})
}.
\label{eq:grpo-advantage}
\end{equation}
The token-level importance ratio between the current and old policies is
\begin{equation}
r_{p,j,t}(\theta)
=
\frac{
\pi_{\theta}(y_{p,j,t}\mid p,y_{p,j,<t})
}{
\pi_{\theta_{\mathrm{old}}}(y_{p,j,t}\mid p,y_{p,j,<t})
}.
\end{equation}
The policy is optimized using the clipped objective
\begin{equation}
\begin{aligned}
\mathcal{J}_{\mathrm{GRPO}}(\theta)
=
\mathbb{E}\Bigg[
\frac{1}{G}\sum_{j=1}^{G}
\frac{1}{|y_{p,j}|}
\sum_{t=1}^{|y_{p,j}|}
\Big[
&\min\big(
r_{p,j,t}(\theta)A_{p,j},\\
\operatorname{clip}(r_{p,j,t}(\theta),1-\delta,1+\delta)A_{p,j}
\big)
&-\beta D_{\mathrm{KL}}^{p,j,t}
\Big]
\Bigg],
\end{aligned}
\label{eq:grpo-objective}
\end{equation}
Here, $r_{p,j,t}(\theta)$ measures the token-level probability change from the old policy to the current policy. Clipping restricts this ratio to $[1-\delta,1+\delta]$, and the minimum forms a conservative PPO-style surrogate that limits overly large policy updates. The factor $1/|y_{p,j}|$ normalizes contributions across responses of different lengths. The KL term regularizes the updated policy toward the reference policy. In our experiments, we set $\beta=0$ and disable both the KL loss in the policy objective and the KL penalty in the reward.

\subsection{Algorithmic Details of LEEPS}
\label{appendix:leeps-algorithm}

Algorithm~\ref{alg:leeps} makes the operational sequence of LEEPS explicit. Before RL training, the sampler extracts a representation for every training prompt and precomputes its latent-neighbor cache. The first $T_{\mathrm{cold}}$ training steps use only the explore portfolio to collect initial rollout feedback. After this cold-start phase, LEEPS estimates the recent non-trivial ratios of the explore and exploit portfolios and adjusts the exploration fraction $\rho$. When either estimate is unavailable, it uses the default exploration fraction $\rho_0$; the resulting value is clipped to $[\rho_{\min},\rho_{\max}]$.

Within the exploit portfolio, prompts are sampled without replacement according to $w_{\mathrm{exploit}}$. Within the explore portfolio, candidates with valid latent-neighbor estimates are ranked by $w_{\mathrm{explore}}$, while candidates without observed neighbors are randomly ordered and used as fallback options. If either portfolio cannot satisfy its assigned quota, the sampler fills the remaining positions from the unselected candidates in the other portfolio. After rollout, $\hat{a}_p$ is replaced by the latest group success rate, $n_p$ is incremented, and the portfolio-level histories are updated. Thus, all online decisions depend only on previously observed outcomes and require no additional selection rollouts.

\begin{algorithm}[H]
\caption{LEEPS training procedure}
\label{alg:leeps}
\textbf{Input}: Training set $\mathcal{D}$, policy $\pi_\theta$, batch size $B$, group size $G$\\
\textbf{Parameters}: Default exploration fraction $\rho_0$, target non-trivial ratio $\tau$, cold-start length $T_{\mathrm{cold}}$, bounds $[\rho_{\min},\rho_{\max}]$
\begin{algorithmic}[1]
\STATE Precompute the latent neighborhood $\mathcal{N}(p)$ for each $p\in\mathcal{D}$.
\STATE Initialize prompt statistics $\{\hat{a}_p,n_p\}$ and portfolio-level histories.
\FOR{training step $t=1,2,\ldots$}
    \STATE Draw a candidate batch $\mathcal{B}_{\mathrm{cand}}$ with $|\mathcal{B}_{\mathrm{cand}}|>B$.
    \STATE $\mathcal{P}_{\mathrm{exploit}}\gets\{p\in\mathcal{B}_{\mathrm{cand}}:n_p>0,\ 0<\hat{a}_p<1\}$.
    \STATE $\mathcal{P}_{\mathrm{explore}}\gets\mathcal{B}_{\mathrm{cand}}\setminus\mathcal{P}_{\mathrm{exploit}}$.
    \IF{$t\le T_{\mathrm{cold}}$}
        \STATE $\rho\gets 1$.
    \ELSE
        \STATE Estimate $q_{\mathrm{explore}}$ and $q_{\mathrm{exploit}}$ from recent rollout outcomes.
        \STATE Choose $\rho$ to target $\tau=\rho q_{\mathrm{explore}}+(1-\rho)q_{\mathrm{exploit}}$; use $\rho_0$ if feedback is insufficient.
        \STATE $\rho\gets\operatorname{clip}(\rho,\rho_{\min},\rho_{\max})$.
    \ENDIF
    \STATE $b_{\mathrm{exploit}}\gets\operatorname{round}((1-\rho)B)$ and $b_{\mathrm{explore}}\gets \operatorname{round}(\rho B)$.
    \STATE Select up to $b_{\mathrm{exploit}}$ prompts from $\mathcal{P}_{\mathrm{exploit}}$ using $w_{\mathrm{exploit}}$.
    \STATE Estimate $\hat{a}^{\mathrm{lat}}_p$ for $p\in\mathcal{P}_{\mathrm{explore}}$ from observed neighbors.
    \STATE Rank scored explore prompts by $w_{\mathrm{explore}}$ and randomly order unscored prompts.
    \STATE Select up to $b_{\mathrm{explore}}$ prompts from $\mathcal{P}_{\mathrm{explore}}$ and backfill any quota shortfall.
    \STATE Combine both selections into $\mathcal{B}_{\mathrm{train}}$ with $|\mathcal{B}_{\mathrm{train}}|=B$.
    \STATE Generate $G$ responses per selected prompt and compute each success rate $a(p)$.
    \STATE Update $\{\hat{a}_p,n_p\}$ and the portfolio-level histories using the observed outcomes.
    \STATE Update $\pi_\theta$ with GRPO using the generated responses.
\ENDFOR
\STATE \textbf{return} $\pi_\theta$
\end{algorithmic}
\end{algorithm}

\paragraph{Complexity and Scalability.}
LEEPS introduces two sources of additional cost beyond standard RLVR training: one-time initialization and per-step prompt selection. Let $N$ denote the number of training prompts, $d$ the dimension of each prompt representation, $M=|\mathcal{B}_{\mathrm{cand}}|$ the candidate-batch size, $K_{\mathrm{stored}}$ the number of nearest neighbors stored for each prompt, and $K\leq K_{\mathrm{stored}}$ the maximum number of neighbors with observed rollout outcomes used for success-rate estimation. During initialization, extracting all prompt representations requires $O(NC_{\mathrm{enc}})$ time, where $C_{\mathrm{enc}}$ is the cost of encoding one prompt with the backbone model, and constructing the exact nearest-neighbor lists requires $O(N^2d)$ time. The resulting representations, neighbor indices, and similarities require $O(Nd+NK_{\mathrm{stored}})$ persistent storage. During RL training, aggregating up to $K$ observed neighbors for each of the $M$ candidate prompts gives the $O(MK)$ scoring cost, while sorting at most $M$ exploration candidates by their scores gives the $O(M\log M)$ ranking cost. If the stored lists must be fully scanned to find observed neighbors, the scoring cost is at most $O(MK_{\mathrm{stored}})$. The online sampler also stores the two rollout-dependent statistics $\hat{a}_p$ and $n_p$ for each prompt, requiring $O(N)$ storage, and temporarily stores $O(M)$ candidate scores at each step. Consequently, when $M$, $K$, and $K_{\mathrm{stored}}$ are fixed, increasing the training-set size $N$ mainly increases initialization time and persistent storage, without directly increasing the per-step selection time. Similarly, increasing the backbone size mainly raises $C_{\mathrm{enc}}$ and typically $d$, again affecting initialization rather than online selection, which requires no additional model forward passes. 

\section{Experimental Details}
\label{appendix:setup}

\subsection{Training Data}

All methods are trained on DAPO-Math-17K~\cite{yu2026dapo}, which provides mathematical reasoning problems paired with reference answers for rule-based verification. We use the original problem text without additional rewriting. For both RLVR training and mathematical reasoning evaluation, each problem is represented by the following messages and then serialized using the model's native chat template:

\begin{center}
\begingroup
\setlength{\fboxsep}{6pt}
\setlength{\fboxrule}{0.5pt}
\fbox{%
\begin{minipage}{0.90\columnwidth}
\small\ttfamily\raggedright
\textbf{System:} Please reason step by step, and put your final answer within \textbackslash boxed\{\}.\\[3pt]
\textbf{User:} \{question\}
\end{minipage}%
}
\endgroup
\end{center}

During RLVR training, a rule-based verifier extracts the final boxed answer and checks it against the reference answer to produce a binary reward.

\subsection{Training Configuration}

Table~\ref{tab:training-hyperparameters} summarizes the main training hyperparameters. All methods at both model scales share the common settings. The LEEPS-specific settings control candidate construction, adaptive explore--exploit allocation, and latent-neighbor retrieval. Following DAPO~\cite{yu2026dapo}, we adopt the Clip-Higher strategy, setting the lower and upper policy-ratio clipping coefficients to 0.20 and 0.28, respectively, to permit larger probability increases and mitigate premature entropy collapse. We disable both the KL loss in the policy objective and the KL penalty in the reward, following the prior work~\cite{qu2026can,mao2026dynamics}.

\begin{table}[t]
\centering
\small
\setlength{\tabcolsep}{2pt}
\begin{tabular}{@{}p{0.62\columnwidth}cc@{}}
\toprule
\textbf{Hyperparameter} & \textbf{1.5B} & \textbf{7B} \\
\midrule
\multicolumn{3}{c}{\textit{Common Settings}} \\
\addlinespace[2pt]
Prompt batch size $B$ & 256 & 256 \\
PPO mini-batch size & 64 & 64 \\
Rollouts per prompt $G$ & 8 & 8 \\
Maximum prompt length & 1024 & 1024 \\
Maximum response length & 3072 & 3072 \\
Learning rate & $1\times10^{-6}$ & $1\times10^{-6}$ \\
Weight decay & 0.1 & 0.1 \\
AdamW $\beta_1/\beta_2$ & 0.9/0.999 & 0.9/0.999 \\
Rollout temperature & 1.0 & 1.0 \\
Top-$p$ & 1.0 & 1.0 \\
Clipping range (low/high) & 0.20/0.28 & 0.20/0.28 \\
Entropy coefficient & 0.001 & 0.001 \\
Gradient clipping & 1.0 & 1.0 \\
KL loss & Disabled & Disabled \\
KL reward penalty & Disabled & Disabled \\
\midrule
\multicolumn{3}{c}{\textit{LEEPS-Specific Settings}} \\
\addlinespace[2pt]
Candidate ratio $|\mathcal{B}_{\mathrm{cand}}|/B$ & 16 & 16 \\
Default exploration ratio $\rho_0$ & 0.15 & 0.15 \\
Cold-start steps $T_{\mathrm{cold}}$ & 5 & 5 \\
Target non-trivial ratio $\tau$ & 0.90 & 0.90 \\
Adaptive window & 3 & 3 \\
Exploration-ratio range $[\rho_{\min},\rho_{\max}]$ & 0.05--0.95 & 0.05--0.95 \\
Number of nearest neighbors $K$ & 64 & 64 \\
Cached candidate neighbors & 1024 & 1024 \\
Embedding layer & 22 & 26 \\
\bottomrule
\end{tabular}
\caption{Training hyperparameters for Qwen2.5-Math-1.5B and Qwen2.5-Math-7B. Common settings are shared by all methods, while the lower block contains parameters specific to LEEPS.}
\label{tab:training-hyperparameters}
\end{table}

For latent-neighbor construction, we use the final-token hidden states from layer 22 of Qwen2.5-Math-1.5B and layer 26 of Qwen2.5-Math-7B. For each prompt, we cache its 1,024 nearest candidates and use up to $K=64$ neighbors with observed rollout outcomes to compute $\hat{a}^{\mathrm{lat}}_p$. The prompt representations and the resulting neighbor cache are computed once before RL training and kept fixed thereafter. During training, LEEPS updates only the rollout-dependent prompt statistics used with this cache. This implementation avoids repeated representation extraction and nearest-neighbor construction while allowing the exploration scores to adapt to newly observed rollout outcomes.

For all offline latent-neighbor AUROC analyses, we use DAPO-Math-17K, the same prompt pool used for subsequent RL training, with a random 75\%/25\% reference--query split (seed 42). Neighbor estimates are constructed from the reference pool and evaluated on the query set; this split is used only for the diagnostic, while RL training uses the complete dataset.

\subsection{Evaluation Benchmarks and Protocol}

\paragraph{Mathematical Reasoning Evaluation.}
The six in-domain benchmarks cover complementary forms and levels of mathematical reasoning:
\begin{itemize}
    \setlength{\itemsep}{2pt}
    \setlength{\topsep}{3pt}
    \item \textbf{MATH-500}~\cite{hendrycks2021measuring} is a curated 500-problem subset of MATH covering algebra, geometry, number theory, counting and probability, and other competition-mathematics subjects.
    \item \textbf{Minerva-Math}~\cite{lewkowycz2022solving} contains 272 quantitative reasoning problems drawn from mathematical and scientific contexts.
    \item \textbf{OlympiadBench}~\cite{he2024olympiadbench} targets challenging olympiad-level reasoning. We use 581 single-answer problems from its English text-only competition-mathematics subset.
    \item \textbf{AMC23}~\cite{li2024numinamath} contains 40 problems from the 2023 American Mathematics Competitions.
    \item \textbf{AIME24}~\cite{li2024numinamath} contains 30 problems from the 2024 American Invitational Mathematics Examination.
    \item \textbf{AIME25}~\cite{li2024numinamath} contains 30 problems from the 2025 American Invitational Mathematics Examination.
\end{itemize}

We use the same message format as in training, with the original problem placed in the user message and the complete messages serialized using the model's native chat template. We report greedy pass@1 with temperature 0 for MATH-500, Minerva-Math, and OlympiadBench. For AMC23, AIME24, and AIME25, we sample 16 responses per problem with temperature 0.6 and top-$p$ 1.0 and report avg@16. The maximum generation length is 3072 tokens in all cases. Generated answers are scored using the same rule-based verification procedure, and the Overall score is the unweighted mean of the six primary benchmark metrics.

\paragraph{OOD General-Reasoning Evaluation.}
We additionally evaluate cross-domain generalization on three multiple-choice benchmarks that are not used for training. Each method is evaluated on the OOD benchmarks using the same checkpoint as for the main mathematical reasoning results.
\begin{itemize}
    \setlength{\itemsep}{2pt}
    \setlength{\topsep}{3pt}
    \item \textbf{GPQA-Diamond}~\cite{rein2023gpqa} contains 198 expert-validated, graduate-level questions in biology, physics, and chemistry.
    \item \textbf{ARC-Challenge}~\cite{allenai:arc} contains 1,172 challenging grade-school science questions that require both scientific knowledge and reasoning.
    \item \textbf{MMLU-Pro}~\cite{wang2024mmlu} contains 12,032 questions spanning a broad range of academic and professional disciplines. Its more challenging questions and expanded candidate sets make it more difficult than the original MMLU benchmark.
\end{itemize}

For OOD evaluation, the original question and its letter-labeled choices are placed in the user message. Because these benchmarks require a choice rather than a free-form mathematical answer, we use the following adjusted system instruction:

\begin{center}
\begingroup
\setlength{\fboxsep}{6pt}
\setlength{\fboxrule}{0.5pt}
\fbox{%
\begin{minipage}{0.90\columnwidth}
\small\ttfamily\raggedright
\textbf{System:} Please reason step by step, and only put your final choice within \textbackslash boxed\{\}.\\[3pt]
\textbf{User:} \{question and letter-labeled choices\}
\end{minipage}%
}
\endgroup
\end{center}

All OOD results are greedy pass@1 accuracies. We use temperature 0, generate one response per question, and set the maximum generation length to 3072 tokens. The predicted choice is extracted from the final boxed expression and compared with the reference option.

\section{Additional Experimental Results}

\subsection{Experiments with the Llama Model}

\paragraph{Experimental Configuration.}
We repeat the main experimental protocol using Llama-3.2-3B-Instruct as the backbone. All settings follow the main experiments unless specified otherwise. The maximum prompt and response lengths are set to 2,048 and 8,192 tokens, respectively. For LEEPS, the target non-trivial ratio is fixed to $\tau=0.8$ before training, and embedding layer 21 is used for the latent-neighbor representation. We evaluate every 10 training steps across the six benchmarks.

\begin{table*}[t]
\centering
\small
\begin{tabular*}{\textwidth}{@{\extracolsep{\fill}} l c c c c c c c c @{}}
\toprule
\textbf{Method} & \textbf{MATH-500} & \textbf{Minerva} & \textbf{Olympiad} & \textbf{AMC23} & \textbf{AIME24} & \textbf{AIME25} & \textbf{Average} & \textbf{Rollouts} $\downarrow$ \\
\midrule
Base & 44.00 & 19.49 & 15.15 & 24.84 & 5.62 & 0.62 & 18.29 & -- \\
GRPO & 49.00 & 19.85 & 19.79 & 34.53 & 11.67 & \textbf{0.83} & 22.61 & \textbf{328K} \\
DS & \textbf{52.00} & \textbf{22.43} & 20.31 & \textbf{40.62} & 9.58 & \underline{0.42} & \underline{24.23} & 1231K \\
MoPPS & 49.80 & 20.96 & \underline{22.03} & 33.12 & \underline{12.08} & \underline{0.42} & 23.07 & \textbf{328K} \\
DPS & 49.20 & \textbf{22.43} & 19.10 & 34.69 & 11.88 & 0.00 & 22.88 & \textbf{328K} \\
LEEPS & \underline{50.80} & \underline{21.32} & \textbf{23.24} & \underline{36.88} & \textbf{13.12} & 0.21 & \textbf{24.26} & \textbf{328K} \\
\bottomrule
\end{tabular*}
\caption{Results on Llama-3.2-3B-Instruct. For each method, we report the checkpoint with the highest average score within the first 160 training steps. The Rollouts column gives the cumulative number of generated responses up to the common 160-step cutoff; DS may generate multiple candidate batches per policy update. Best and second-best scores among the trained methods are shown in \textbf{bold} and \underline{underlined}, respectively.}
\label{tab:llama3b-results}
\end{table*}

\begin{figure*}[t]
\centering
\includegraphics[width=\textwidth]{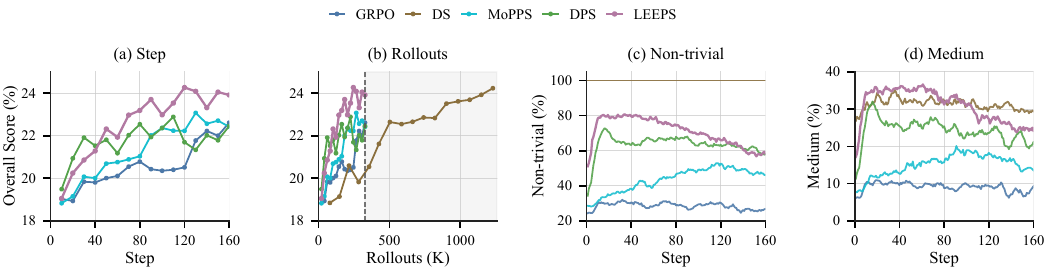}
\caption{Additional results on Llama-3.2-3B-Instruct. Panels (a)--(b) show the overall score versus training step and cumulative rollouts, respectively. DS is omitted from Panel (a) because it may generate multiple candidate batches per policy update, but is included in the rollout-normalized comparison in Panel (b). The dashed line marks the shared 328K-rollout budget of GRPO and the pre-rollout selectors, and the shaded region indicates additional rollouts used by DS. Panels (c)--(d) show the non-trivial ratio and medium ratio of the selected prompt groups, using a 7-step centered moving average.}
\label{fig:llama3b-results-diagnostics}
\end{figure*}

\paragraph{Results and Analysis.}
As shown in Table~\ref{tab:llama3b-results}, LEEPS achieves the highest average score of 24.26. Among the methods using the shared 328K-rollout budget, the strongest baseline is MoPPS at 23.07; LEEPS improves upon it by 1.19 points, corresponding to a relative gain of approximately 5.2\%. DS also performs strongly, reaching 24.23, but uses 1,231K rollouts by the 160-step cutoff, or $3.76\times$ the shared budget. Thus, LEEPS achieves performance comparable to DS while requiring substantially fewer rollouts. Figure~\ref{fig:llama3b-results-diagnostics}(a)--(b) further shows that LEEPS achieves the highest peak among the pre-rollout selectors within the common rollout budget.

The sample characteristics in Figure~\ref{fig:llama3b-results-diagnostics}(c)--(d) help explain this efficiency difference. DS keeps its non-trivial ratio close to 100\% through post-rollout filtering, but this requires generating multiple candidate batches and discarding uninformative responses. LEEPS instead maintains relatively high non-trivial and medium ratios through most of training without generating additional rollouts for selection. GRPO remains lower on both measures; MoPPS has a lower non-trivial ratio, especially early in training; and DPS preserves a relatively high non-trivial ratio but includes fewer medium-difficulty prompts. This pattern suggests that LEEPS retains a favorable balance between sample quality and rollout cost on the Llama backbone, providing further evidence that its sampling strategy extends beyond the Qwen2.5-Math family.

\subsection{Hyperparameter Sensitivity}
We examine the robustness of LEEPS along three dimensions: the amount of rollout-labeled data used for layer calibration, the number of latent neighbors $K$, and the target non-trivial ratio $\tau$. The first two concern the reliability and sample efficiency of the offline latent-neighbor construction, whereas $\tau$ controls the online explore--exploit allocation. The representation layer is selected once before training, using a predefined offline AUROC criterion applied across all candidate layers, rather than being tuned to downstream performance. Because this layer-wise diagnostic directly evaluates the signal used by latent-guided exploration, we focus on the calibration data needed to make a stable layer choice, rather than repeating the full RL procedure for each layer.
\label{appendix:hyperparameter-sensitivity}

\begin{figure}[htbp]
\centering
\includegraphics[width=\linewidth]{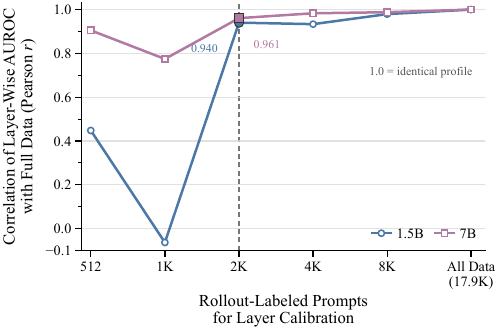}
\caption{Sample efficiency of layer calibration. Pearson $r$ compares the layer-wise AUROC values estimated from each subset with those obtained using all 17,917 prompts; $r=1$ indicates the same cross-layer pattern. Representative calibration sizes are shown, and the vertical line marks 2,048 prompts.}
\label{fig:layer-calibration-sample-size}
\end{figure}

\paragraph{Calibration Sample Size.}
The full-data diagnostic shows that the non-triviality signal strengthens with depth but forms a broad upper-layer plateau, where adjacent layers differ little in AUROC. Layer calibration therefore needs only to locate this near-optimal region rather than recover a unique maximizer. We recompute the diagnostic on nested subsets and measure the Pearson correlation between each subset's hidden-layer AUROC profile (L1--L28) and the full-data profile. A high correlation means that layers relatively stronger or weaker in the full-data profile show similar relative deviations under the subset; it measures preservation of the cross-layer structure rather than equality of absolute AUROCs or exact layer ranks. As shown in Figure~\ref{fig:layer-calibration-sample-size}, the agreement fluctuates for very small subsets, rises sharply by 2,048 prompts, and remains stable thereafter. With 2,048 prompts (11.4\% of the complete pool), the correlations reach 0.940 and 0.961 for the 1.5B and 7B models, while the selected layers remain within 0.005 AUROC of the corresponding full-data optima. Thus, 2,048 rollout-labeled prompts suffice in this diagnostic to make a near-optimal layer choice. The main experiments use the full-data maxima (L22 and L26), fixed throughout RL training.

\begin{figure}[htbp]
\centering
\includegraphics[width=\linewidth]{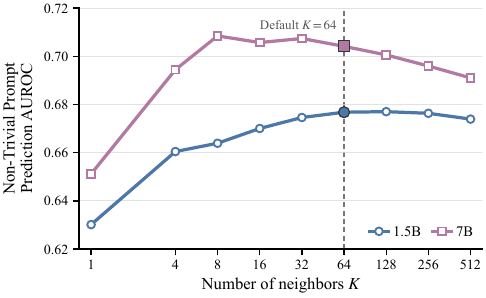}
\caption{Sensitivity of non-trivial-prompt prediction AUROC to the number of nearest neighbors $K$ for Qwen2.5-Math-1.5B and Qwen2.5-Math-7B on DAPO-Math-17K. For each $K$, the representation layer with the highest AUROC is selected independently, so the curves form exploratory best-layer envelopes rather than fixed-layer comparisons. The dashed vertical line marks the default setting $K=64$.}
\label{fig:knn-k-sensitivity}
\end{figure}

\paragraph{Number of Neighbors $K$.}
Using the same offline AUROC setup, we vary $K$ to assess whether the latent-neighbor signal depends critically on neighborhood size and whether a single shared setting can be used across model scales. As shown in Figure~\ref{fig:knn-k-sensitivity}, AUROC improves as the neighborhood expands beyond very small values of $K$, remains stable over a broad intermediate range, and declines mildly for overly large neighborhoods, particularly for the 7B model. This trend is consistent with a locality--stability trade-off: very small neighborhoods provide limited evidence for the estimate, whereas excessively large neighborhoods may dilute local information with less related prompts. Within this exploratory diagnostic, the broad plateau indicates that the latent-neighbor signal is not highly sensitive to the exact choice of $K$ and supports $K=64$ as a robust shared default without per-model tuning.

\paragraph{Target Non-Trivial Ratio $\tau$.}
We vary $\tau\in\{0.80,0.85,0.90,0.95\}$ to assess sensitivity to the shared main-experiment setting $\tau=0.90$. Across the tested range, all settings yield broadly competitive Overall Score trajectories at both model scales, indicating that LEEPS is not highly sensitive to the exact choice of $\tau$. Increasing $\tau$ from $0.80$ to $0.90$ raises the realized non-trivial ratio, whereas $\tau=0.95$ provides little further increase and can reduce performance. Although $\tau=0.90$ attains the highest peak scores, the other settings remain competitive, suggesting a broad effective range rather than a narrowly tuned optimum. Overall, $\tau$ controls the balance between exploiting known non-trivial prompts and exploring uncertain prompts.

\begin{figure*}[htbp]
\centering
\includegraphics[width=\textwidth]{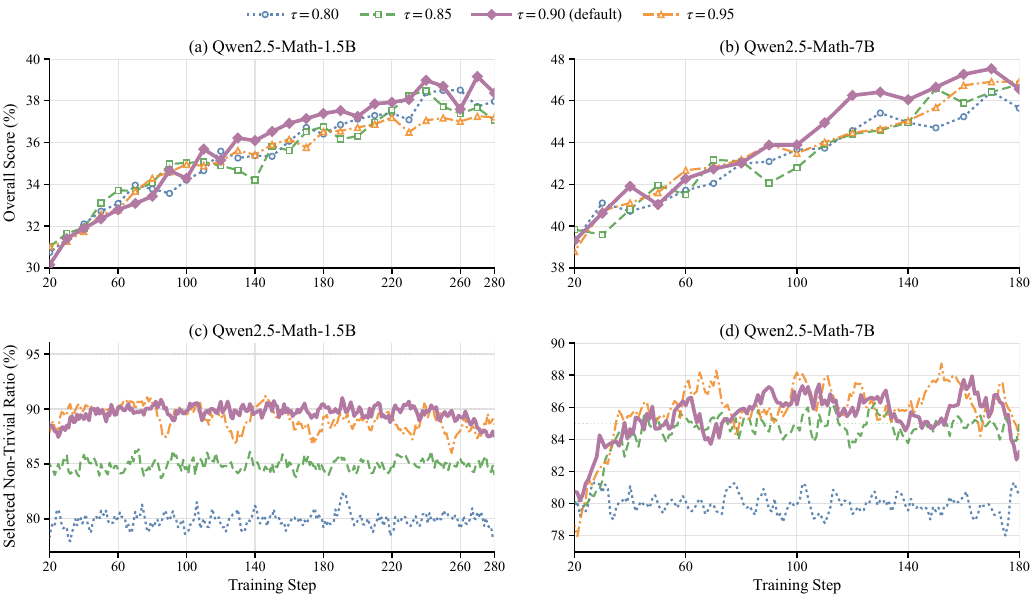}
\caption{Sensitivity of LEEPS to the target non-trivial ratio $\tau$ on Qwen2.5-Math-1.5B and Qwen2.5-Math-7B. Panels (a)--(b) show the Overall Score, while Panels (c)--(d) show the selected-batch non-trivial ratio using a 7-step centered moving average. The default $\tau=0.90$ is highlighted, and all curves are shown from training step 20 for clarity.}
\label{fig:target-ratio-sensitivity}
\end{figure*}

\begin{figure*}[htbp]
\centering
\includegraphics[width=\textwidth]{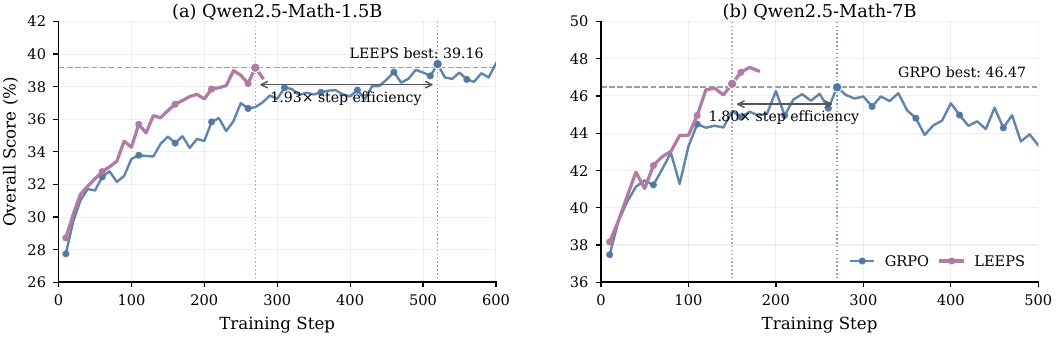}
\caption{Extended training-step comparison between GRPO and LEEPS. LEEPS is shown through the model-specific ranges used in the main experiments (280 steps for 1.5B and 180 steps for 7B), whereas GRPO is extended to 600 and 500 steps, respectively. Dashed horizontal lines indicate a common performance target: the best LEEPS score for the 1.5B model and the best GRPO score for the 7B model. Dotted vertical lines mark the first step at which each method reaches the target. LEEPS achieves $1.93\times$ and $1.80\times$ higher training-step efficiency on the 1.5B and 7B models, respectively.}
\label{fig:grpo-efficiency-extended}
\end{figure*}

\subsection{Extended Training Efficiency against GRPO}

LEEPS achieves $1.93\times$ and $1.80\times$ higher training-step efficiency than GRPO on the 1.5B and 7B models, respectively, as shown in Figure~\ref{fig:grpo-efficiency-extended}. On the 1.5B model, LEEPS reaches an overall score of 39.16 at step 270, whereas GRPO requires 520 steps to reach the same level. On the 7B model, LEEPS surpasses the best GRPO score of 46.47 at step 150, while GRPO reaches this score at step 270. The extended trajectories further show that GRPO improves only gradually at the later stage on the 1.5B model, while its 7B performance plateaus around 45--46 and declines after reaching its peak. Because GRPO and LEEPS generate the same number of rollouts per training step, the efficiency factors also apply to their rollout budgets. These observations indicate that LEEPS reaches strong performance earlier through more effective prompt selection.

\section{Data Examples}
\label{appendix:data-examples}

Below, we present one DAPO-Math-17K training example and one example from each OOD benchmark. The OOD examples reproduce the normalized choice order used for evaluation. These benchmarks already provide answer candidates: GPQA-Diamond supplies one correct answer and three distractors, ARC-Challenge supplies labeled choices, and MMLU-Pro supplies an option list. Thus, no open-ended question is converted into multiple-choice form.

\newcommand{\dataexamplebox}[2]{%
\par\medskip
\noindent
\begingroup
\setlength{\tabcolsep}{6pt}%
\renewcommand{\arraystretch}{1.15}%
\begin{tabular}{|p{\dimexpr\columnwidth-2\tabcolsep-2\arrayrulewidth\relax}|}
\hline
\textbf{#1 Data Example} \\
\hline
\small #2 \\
\hline
\end{tabular}%
\endgroup
\par
}

\dataexamplebox{DAPO-Math-17K}{%
\textbf{Question:}\par
What is the tens digit of $5^{2005}$?
\par\smallskip
\textbf{Reference Answer:} $2$.
}

\dataexamplebox{GPQA-Diamond}{%
\textbf{Question:}\par
Which of the following (effective) particles is not associated with a spontaneously-broken symmetry?
\par\smallskip
\textbf{Choices:}\par
\begin{tabular*}{\linewidth}{@{\extracolsep{\fill}}ll@{}}
(A) Phonon & (B) Skyrmion \\
(C) Pion   & (D) Magnon
\end{tabular*}
\par\smallskip
\textbf{Answer:} (B) Skyrmion.
}

\dataexamplebox{ARC-Challenge}{%
\textbf{Question:}\par
An astronomer observes that a planet rotates faster after a meteorite impact. Which is the most likely effect of this increase in rotation?
\par\smallskip
\textbf{Choices:}\par
\begin{tabular}{@{}p{0.48\linewidth}@{\hspace{4pt}}p{0.48\linewidth}@{}}
(A) Planetary density will decrease. & (B) Planetary years will become longer. \\
(C) Planetary days will become shorter. & (D) Planetary gravity will become stronger.
\end{tabular}
\par\smallskip
\textbf{Answer:} (C) Planetary days will become shorter.
}

\dataexamplebox{MMLU-Pro}{%
\textbf{Question:}\par
The symbol for antimony is
\par\smallskip
\textbf{Choices:}\par
\begin{tabular*}{\linewidth}{@{\extracolsep{\fill}}lllll@{}}
(A) Am & (B) As & (C) Ab & (D) Au & (E) Fe \\
(F) Ag & (G) An & (H) At & (I) W  & (J) Sb
\end{tabular*}
\par\smallskip
\textbf{Answer:} (J) Sb.
}

\end{document}